\documentclass[11pt]{article}

\usepackage[]{acl}

\usepackage{hyperref}       
\usepackage{times}
\usepackage{latexsym}

\usepackage[T1]{fontenc}

\usepackage[utf8]{inputenc}

\usepackage{microtype}

\usepackage{subcaption}
\usepackage{booktabs}       
\usepackage{amsfonts}       
\usepackage{nicefrac}       
\usepackage{multirow}
\usepackage{listings}
\usepackage{algorithm}
\usepackage{enumitem}
\usepackage{amsmath,amssymb}
\usepackage{pifont}
\usepackage{color, colortbl}
\usepackage{caption}
\usepackage{array}
\usepackage{comment}
\usepackage{todonotes}
\usepackage{varwidth}
\usepackage{colortbl}

\lstdefinestyle{examples}{
    frame=single,
    framerule=0pt,
    basicstyle=\ttfamily\small,
    backgroundcolor=\color{gray!10!white},
    columns=flexible,
    breaklines=false,
    keepspaces=false,
}

\lstdefinestyle{python}{
    frame=single,
    framerule=0pt,
    language=Python,
    basicstyle=\ttfamily\small,
    keywordstyle=\bfseries,
    backgroundcolor=\color{gray!10!white},
    commentstyle=\itshape\color{blue!60},
    columns=flexible,
    breaklines=true,
    keepspaces=true,
    showspaces=false
}

\lstdefinestyle{custom}{
    otherkeywords={[comments],[python],[eoc],[descr],[sos],[new\_line],[indent],[dedent]},
    frame=single,
    framerule=0pt,
    basicstyle=\ttfamily\small,
    keywordstyle=\bfseries\color{blue!60},
    backgroundcolor=\color{gray!10!white},
    keepspaces=true,
    breaklines=true,
    tabsize=4,
    columns=flexible,
    breakindent=0pt,
    numbers=none,
    tab=\t,
    showspaces=false, 
}

\title{Text-to-Code Generation with Modality-relative Pre-training}

\author{
Fenia Christopoulou, Guchun Zhang, Gerasimos Lampouras \\
Huawei Noah's Ark Lab, London, UK \\
\small \texttt{\{efstathia.christopoulou, guchun.zhang, gerasimos.lampouras\}@huawei.com} \\
}

\begin{document}
\maketitle

\begin{abstract}
Large pre-trained language models have recently been expanded and applied to programming language tasks with great success, often through further pre-training of a strictly-natural language model--where training sequences typically contain both natural and (linearised) programming language. 
Such approaches effectively map both modalities of the sequence into the same embedding space. However, programming language keywords (e.g. ``while'') often have very strictly defined semantics. As such, transfer learning from their natural language usage may not necessarily be beneficial to their code application and vise versa. Assuming an already pre-trained language model, in this work we investigate how sequence tokens can be adapted and represented differently, depending on which modality they belong to, and to the ultimate benefit of the downstream task. 
We experiment with separating embedding spaces between modalities during further model pre-training with modality-relative training objectives.
We focus on text-to-code generation and observe consistent improvements across two backbone models and two test sets, measuring pass@$k$ and a novel incremental variation.\footnote{Code, data and models are publicly available at \url{https://github.com/huawei-noah/noah-research/tree/master/NLP/text2code_mrpt}}
\end{abstract}

\section{Introduction}

Increasingly more large pre-trained language models~\cite[PLMs]{radford2018improving,devlin-etal-2019-bert,raffel-etal-2020-exploring,han-etal-2021-models} based on the Transformer~\cite{vaswani2017attention} architecture have been proposed and shown to achieve state-of-the-art results on a variety of Natural Language Processing (NLP) tasks.
Lately, such models have been adapted to more specific language domains, e.g. Biomedical~\cite{huang-etal-2019-clinicalbert,lee2020biobert,DBLP:journals/corr/abs-1903-10676}, Legal~\cite{chalkidis-etal-2020-legal}, Cyber Security~\cite{https://doi.org/10.48550/arxiv.2204.02685} and Finance~\cite{araci-2019-finbert}, while also expanding to include signals from modalities other than natural language, such as Vision~\cite{DBLP:journals/corr/abs-1906-02940,chen2022when,pmlr-v119-chen20s,carion2020end,touvron-etal-2021-training,dosovitskiy2021an}, Proteins~\cite{brandes2022proteinbert}, Time Series~\cite{DBLP:journals/corr/abs-2001-08317, 10.1145/3394486.3403118,qin2022tsbert} and Code~\cite{kanade2020pretrained,wang-etal-2021-codet5,feng-etal-2020-codebert,guo-etal-2022-unixcoder,phan-etal-2021-cotext}.

\begin{figure}[t!]
    \centering
    \includegraphics[width=\linewidth]{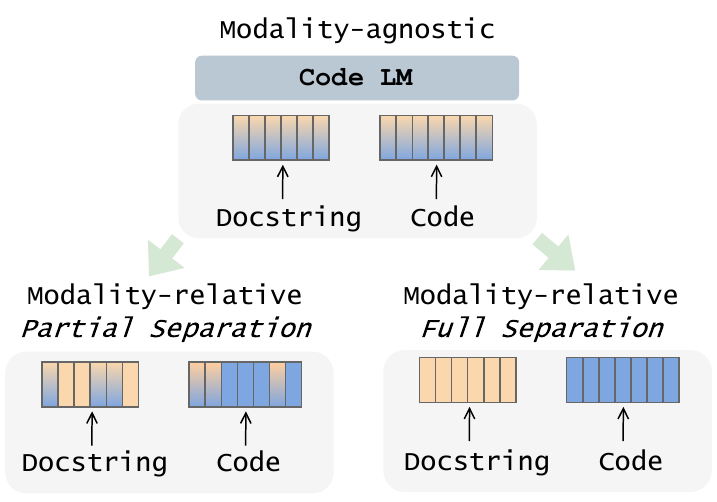}
    \caption{Overview of modality-agnostic/relative pre-training and training objectives.}
    \label{fig:big_picture}
\vspace{-3mm}
\end{figure}

In this work, we experiment with pre-trained language models specifically created for text-to-code generation, i.e. the task of program synthesis given Natural Language (NL) descriptions (e.g. problem definitions or docstrings). While some of the proposed models for this task adopt the encoder-decoder architecture~\cite{alpha_code}, most are trained as decoder-only Transformer models~\cite{codex,fried2022incoder,nijkamp2023codegen,li2023starcoder,roziere2023code}. The latter approach is often preferred, since single-component architectures are well-suited for continuous training over vast amounts of raw non-annotated data, such as Github.\footnote{\url{https://github.com}}

When employing a single-component architecture, all tokens in the sequence are often vectorised via the same embedding layer; effectively assuming that all parts of the sequence share the same semantic space despite language or modality. This may be well suited for many NLP tasks or multilingual models, as transfer learning between languages has been shown to be beneficial~\cite{conneau-etal-2020-unsupervised}. However, we posit that due to their strict technical and logical implications, programming language tokens have distinct semantic meaning that does not necessarily benefit by transfer learning from natural language. To investigate this, we propose modality-relative embedding-spaces and training objectives. Specifically, our work treats the docstring and code subsequences of our data as separate \textit{modalities}, and further tunes separate embeddings with different training objectives--initially tuning them on a shared space (see Figure~\ref{fig:big_picture}).

As our test case, we focused on Python and carried out zero-shot experiments demonstrating that our modality-relative embeddings and training objectives achieve consistent improvements over two baselines on the HumanEval~\cite{codex} and MBPP~\cite{jain-etal-2021-jigsaw} datasets. We also introduce a novel variation of the pass@$k$~\cite{codex} evaluation metric for program synthesis, namely \textit{incremental pass@$k$}, that combines both synthesis and completion tasks to better differentiate models capabilities.

\begin{table}[t!]
\lstset{linewidth=6.1cm}
\centering
\begin{tabular}{>{\footnotesize\sc}p{0.8cm}l}

descr. &
\begin{lstlisting}[style=examples]
Return a greatest common 
divisor of two integers a and b \end{lstlisting} \\ 

sign. &
\begin{lstlisting}[style=python]
def gcd(a: int, b: int) -> int: \end{lstlisting} \\ 

body &  
\begin{lstlisting}[style=python]
while b:
    a, b = b, a % b 
return a \end{lstlisting} \\ 
tests & 
\begin{lstlisting}[style=examples]
gcd(3, 5) = 1  
gcd(25, 15) = 5 \end{lstlisting} \\
    
\end{tabular}
\caption{Example from the HumanEval dataset.}
\label{tab:HE_example}
\vspace{-3mm}
\end{table}

\section{Methodology}

In this work, we focus on the task of text-to-code generation, i.e. to synthesise running code that successfully solves the problem described in a NL description. Table~\ref{tab:HE_example} shows an example from the HumanEval dataset~\cite{codex}, which is typical for the task; the input consists of a NL description of the problem, followed by the function signature of the solution (the function argument types and the expected return type may or may not be included). In addition, the problems may be accompanied by some unit tests, in the form of function calls with specific inputs and the corresponding expected outputs, e.g. \texttt{greatest\_common\_divisor(3, 5) = 1}. The model is then asked to produce the body of the code, where its proper functionality is evaluated against a number of held-out unit tests. 
This task is closely related to code completion, where the input also includes partial code.

\subsection{Modality-agnostic Pre-Training}
\label{sec:stage1_training}

Following previous work~\cite{codex}, here we assume access to already pre-trained language models on raw natural language and code data (namely, Code LMs) as our starting point.
These models have been pre-trained with modality-agnostic pre-training (MAPT), i.e. training that is agnostic with respect to the underlying modalities and all token embeddings are shared between the two.
We employ the PyCodeGPT~\cite{cert2022zan} model that is based on GPT-Neo~\cite{gptneo}, pre-trained on an undisclosed subset of open source Github repositories in Python. 
In addition, we pre-train PanGu-Coder~\cite{christopoulou2022pangu} with regular Causal Language Modelling (CLM)~\cite{radford2018improving} on a different batch of Python data from Github (see Appendix~\ref{sec:data} for details regarding our data collection process).

Our proposed modality-relative training objectives and embedding-spaces are subsequently applied through \textit{further pre-training} of the Code LMs on program synthesis-specific data, consisting solely of text-to-code pairs.
Through this two-stage pre-training approach  
a model is able to learn how to encode both general code structures and natural language through raw data in the first stage, and later focus on how to best generate the correct output code given the NL input.

\subsection{Modality-relative Continual Pre-Training}
\label{sec:stage_2_training}

\begin{figure*}[!ht]
    \centering
    \includegraphics[width=\linewidth]{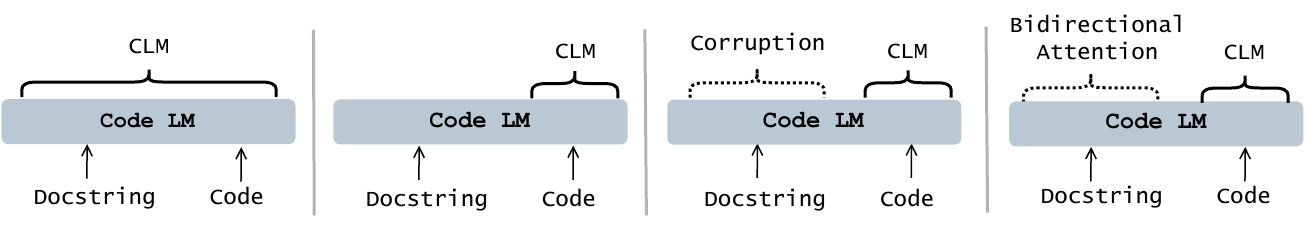}
    \caption{\textsc{text-code clm}, \textsc{code-clm}, \textsc{corrupt-code-clm} and \textsc{prefix-code-clm} pre-training objectives.}
    \label{fig:training_objectives}
\vspace{-3mm}
\end{figure*}

Assuming a modality-agnostic Code LM (e.g. PyCodeGPT or PanGu-Coder) that has attained general knowledge about NL and code during initial pre-training, we continue pre-training with a dedicated focus on the downstream task of text-to-code generation. The training data we use consist of functions crawled from existing, public GitHub repositories. If a function is accompanied by a docstring, we assume that to be a corresponding NL description of the function forming a text-to-code pair instance. 
Details on how the data were gathered and filtered can be found in Appendix~\ref{sec:data}. 

Formally, we denote a training instance as a single sequence of tokens $X = X_D + X_C = \{d_1, ..., d_{N_D}, c_1, ..., c_{N_C}\}$, with $X_D$ and $X_C$ corresponding to all the tokens of the \textit{docstring} modality and the \textit{code} modality of the input, respectively. 
CLM is then applied as follows:
\begin{equation*}
    \mathcal{L}_\textsc{clm}(X) = - \sum_{n=1}^{N} \log p(x_n| x_{<n}; \theta),
    \label{eq:clm}
\end{equation*}
where $N = N_D + N_C$ indicates the total number of tokens in the input sequence $X$ (with $N_D$, $N_C$ the number of \textit{docstring} and \textit{code} tokens, respectively)
and $\theta$ corresponds to the model parameters. 

An important design decision was to formulate the data specifically for function-level code synthesis. As such, we ensured that the training data contain functions that are always accompanied by a natural language description (i.e. using the docstring as the function's problem description), essentially aligning NL with code.
Based on this data format, we propose to train the model via modality-relative pre-training (MRPT), i.e. 
treating the docstring and code subsequences of each instance distinctly.

\subsubsection{Space Separation}
\label{sec:embedding_separation}

To investigate our hypothesis that code tokens should be distinct from NL ones due to the strict semantic meaning of the former, we propose separating the embedding space $E$ after MAPT (see Figure~\ref{fig:big_picture}) through the following two strategies:

\paragraph{Partial Embedding Separation (PES)}
Reasonably, programming language-specific tokens such as \texttt{join}, \texttt{for}, \texttt{return}, \texttt{class}, etc, might suffer more from conflicting signals during modality-agnostic pre-training. On the other hand, tokens used in variables and function names should remain shared between NL and code, as transfer learning can benefit their encoding.
In our first space separation approach, namely \textit{Partial Embedding Separation} (PES), we associate separate embeddings $e_{code}$ and $e_{NL}$ only to tokens that appear in the programming language's grammar. Since we are using Python as our use-case, we extract all necessary tokens from the official Python grammar\footnote{\url{https://docs.python.org/3/reference/grammar.html}} and built-in functions.\footnote{\url{https://docs.python.org/3/library/functions.html}}
Although this method creates two embeddings only for language-related tokens, minimally expanding the model's embedding space, it requires manual effort and is dependent on the programming language one investigates.

\paragraph{Full Embedding Separation (FES)}
Since PES is programming language-specific and cannot capture the distinction among other tokens that might be equally important, our second approach separates the entire embedding space $E$ of the modality-agnostic model, namely \textit{Full Embedding Separation}.
We thus associate a distinct embedding $e_{NL}$ and $e_{code}$ with each modality for the entire vocabulary of the model. This effectively results in doubling the number of embeddings of the model, i.e. $|E_{NL}| + |E_{code}| = 2 * |E|$.

\vspace{0.5em}
\par In both PES and FES, the NL and code embeddings are initialised with their values as they stand after MAPT and are further trained through MRPT.

Alternatively, the separation of embeddings could be performed by using two distinct tokenizers for the NL and code part respectively, but that would forego the MAPT stage and any potential benefit resulting from it. Preliminary experiments were inconclusive on whether using distinct tokenizers was beneficial compared to starting from a modality-agnostic model with a shared embedding space. We opted to keep the MAPT stage constant across all settings, to keep comparisons fair.


\subsubsection{Training Objectives}
\label{sec:training_objectives}

We train modality-aware models with a few training objectives, as shown in Figure~\ref{fig:training_objectives} and detailed below.
As a baseline objective we consider standard causal language modelling over the entire input sequence, which we denote as \textsc{text-code-clm} seen in the leftmost part of Figure~\ref{fig:training_objectives}. These training objectives can then be further combined with the proposed embedding separation strategies.

\paragraph{\textsc{code-clm}: Causal Language Modelling on Code} 

We calculate the loss only on the code subsequence, following \citet{codex}, namely \textsc{code-clm}, training implicitly on the docstring. 
Each code token is predicted based on all previous tokens, including the tokens of the docstring.

\paragraph{\textsc{corrupt-code-clm}: Corrupted Docstring}

We also experiment with corrupting the docstring by randomly masking out some of its tokens.
Specifically, a set of random tokens with indices $X_M = \{\pi_1, ..., \pi_M\}$ in the docstring are replaced with a mask (\texttt{[MASK]}), a random token, or the same token with $0.8$/$0.1$/$0.1$ chance, respectively, similar to \citet{devlin-etal-2019-bert}.
Analogous to the previous objective, we do not measure any loss over the docstring but solely on the code that has to be generated based on the corrupted input.

\paragraph{\textsc{prefix-code-clm}: Bidirectional Attention on Docstring}

Since the down-stream task is not reliant on next word prediction for the docstring, we experiment with allowing bidirectional attention on the docstring tokens, as we assume that additional context can result in better docstring representations. This is similar to how prefix language models work~\citep{dong-etal-2019-unilm,bao-etal-2020-unilmv2,guo-etal-2022-unixcoder}, with the difference that we do not calculate any loss over the prefix, which in our scenario, corresponds to the docstring~\citep{pmlr-v162-wang22u}.
Due to the formatting of their input (see Section~\ref{sec:data_formulation}), in PyCodeGPT we consider both the docstring and the signature as part of the prefix, while in PanGu-Coder only the docstring is playing the role of the prefix.

\section{Experimental Settings}

\subsection{Data Formulation}
\label{sec:data_formulation}

We formulate each model's input according to their original training data format.
PanGu-Coder's input is formed by combining a docstring and its corresponding code as follows, since it was designed to accommodate both code-only and text-code pairs during MAPT pre-training (see Appendix~\ref{app:pangu_mapt}). 
\begin{lstlisting}[style=custom]
[descr] docstring [python] signature code [eoc] 
\end{lstlisting}
where \texttt{[descr]} indicates the beginning of a description, \texttt{[python]} indicates the beginning of the code and \texttt{[eoc]} corresponds to the end of the code sequence.
On the other end, PyCodeGPT was trained on general purpose, raw code repositories, that follow the standard format of the docstring appearing after the code signature, with a start-of-sequence token (\texttt{[sos]}) at the beginning of each instance. We form its input as shown below:
\begin{lstlisting}[style=custom]
[sos] signature [descr] docstring [python] code [eoc]
\end{lstlisting}

\subsection{Inference}
We follow the standard (left-to-right) decoding process used for auto-regressive language models, with temperature scaling ($t$) and nucleus sampling ($p$)~\citep{HoltzmanBDFC20}. Inference adopts a prompt that is similar to the data format used during pre-training, for each model respectively (see Section~\ref{sec:data_formulation}), until the keyword \texttt{[python]}, after which the model is requested to generate the problem solution.
Generation continues until the \texttt{[eoc]} token is generated or a maximum length is reached. 
In the case of the \textsc{prefix-code-clm} objective, we allow bidirectional attention during inference on the given prompt, similar to pre-training.
Due to the nature of our applied tasks, we assume there is always a problem description available during inference. We remove any superfluous white-spaces and line breaks from the descriptions.

Formally, the prompt is a sequence of tokens $P = P_D + P_S = \{d_1, ..., d_{N_D}, s_1, ..., s_{N_S}\}$ where $N_D, N_S$ denote the number of tokens in the docstring and the signature, respectively. In the case of PyCodeGPT the prompt is formatted as $P = P_S + P_D$. The model then generates a continuation $C'$ of the prompt in a left-to-right manner, decoding one token at a time while attending on previous. 
\begin{align*}
C^\prime(P) &= \textsc{PanGu-Coder}(c^\prime_{t}|c^\prime_{<t}, d_{<N_d}, s_{<N_S}) \\   
C^\prime(P) &= \textsc{PyCodeGPT}(c^\prime_{t}|c^\prime_{<t}, s_{<N_S}, d_{<N_D})
\end{align*}

\subsection{Evaluation}
 \label{sec:evaluation}
 
To evaluate our models, we consider two commonly used datasets for checking the functional correctness of generated programs: HumanEval~\cite{codex}, and the Mostly Basic Programming Problems~\cite[MBPP]{austin2021program}. 

HumanEval\footnote{\url{https://github.com/openai/human-eval}} contains 164 handcrafted Python problems accompanied by a set of held-out unit tests (average of 7.7 unit tests per problem), \textit{all of which} must pass in order to count as a successful solution. The dataset prompts include a problem description and a signature. 
In a similar vein, MBPP\footnote{\url{https://github.com/google-research/google-research/tree/master/mbpp}} is comprised of 974 programming problems (374 train, 90 validation, 500 test and 10 few-shot) designed to be solved by entry-level Python programmers. We use only the test set for evaluation.

We checked the overlap of any natural language prompt from all datasets in our training data with exact string matching (non-whitespace text) and found 1 contaminated example with HumanEval and none with the MBPP test set.
Evaluation was performed via adapting the CodeGeeX framework~\cite{codegeex}.\footnote{\url{https://github.com/THUDM/CodeGeeX}}

\subsection{Evaluation Metrics}
\label{sec:eval_metrics}

In order to estimate model performance, we sample $n$ programs/solutions per problem and calculate the unbiased estimator of pass@$k$ for $k$ = [$1$, $10$, $100$], originally introduced in \citet{NEURIPS2019_7298332f} and further adapted by \citet{codex}. 
In all experiments, we use greedy decoding for pass@$1$ and a sample size of $n=200$ with $p=0.8$, $t=0.95$ for $k$ = [$10$, $100$]. Those values were selected via preliminary analysis and remain fixed for all experiments (no further tuning).

\paragraph{\textbf{Incremental pass@$k$}:}

A drawback of pass@$k$ is that it only considers a code solution correct if it passes all provided unit tests, and provides no partial credit for incomplete programs however close they may be to a valid solution. Other metrics may provide partial credit based on syntactic or semantic overlap between the generated program and a reference, e.g. CodeBLUE~\cite{ren2020codebleu} and CodeBERTScore~\cite{zhou2023codebertscore}, but these do not directly check for functional correctness. 

In an effort to provide more granular functional correctness, we propose and report a variation on the pass@$k$ metric; incremental pass@$k$, which aims to measure the code completion capabilities of Code LMs in addition to full program synthesis. 
We automatically create an augmented test set for each dataset by using the provided code solution reference to construct partial code solutions as additional evaluation prompts; partial solutions are constructed by incrementally adding one line of the reference to the previous prompt. 
The pass@$k$ metric is then calculated over the original and augmented prompts, reporting micro-averaged scores.

\subsection{Training details}
We train models using the Adam optimiser~\citep{adam} with $\beta_1 = 0.9$, $\beta_2 = 0.95$ and weight decay of $0.01$. For MAPT, the maximum learning rate is set to $1e^{-4}$, which is decayed by a cosine scheduler until $5e^{-6}$. For MRPT, the maximum and minimum learning rates are $1e^{-5}$ and $5e^{-6}$, respectively. The gradients are clipped at $3.0$ during modality-agnostic and $1.0$ during modality-relative pre-training.
For both models we set the maximum sequence length to $1,024$ tokens, the batch size to $1,024$ instances and warm-up to $1\%$ of the total training steps. 

For MAPT, PanGu-Coder is initialised with random weights and trained for 500K steps from scratch (more details in Appendix~\ref{app:pangu_mapt}). For MRPT, we initialise the models from their MAPT checkpoint and continue training for 5 epochs for PyCodeGPT and 10 epochs for PanGu-Coder.
We report performance of all models at the end of modality-relative pre-training.
Further technical details can be found in Appendix~\ref{app:training_details}.


\begin{table*}[!ht]
\begin{subtable}[b]{\linewidth}
\centering
\scalebox{0.75}{
    \begin{tabular}{lrrrc!{\color{gray!30}\vrule}ccc|ccc!{\color{gray!30}\vrule}ccc}
    \toprule
    & 
    & \multicolumn{3}{c}{\sc{\textbf{human eval}}}
    & \multicolumn{3}{c}{\sc{\textbf{INCR human eval}}}
    & \multicolumn{3}{c}{\sc{\textbf{mbpp}}}  
    & \multicolumn{3}{c}{\sc{\textbf{INCR mbpp}}} \\   
    & \textsc{separation} 
    &  \sc{p@$1$} &  \sc{p@$10$} &  \sc{p@$100$}  
    &  \sc{p@$1$} &  \sc{p@$10$} &  \sc{p@$100$}  
    &  \sc{p@$1$} &  \sc{p@$10$} &  \sc{p@$100$} 
    &  \sc{p@$1$} &  \sc{p@$10$} &  \sc{p@$100$} \\\cmidrule(lr){2-2} \cmidrule(lr){3-5} \cmidrule(lr){6-8} \cmidrule(lr){9-11} \cmidrule(lr){12-14}
           
\textsc{mapt} 
& -
& \cellcolor{lightgray!30}9.15 &	\cellcolor{lightgray!30}13.26 &	\cellcolor{lightgray!30}20.80
& \cellcolor{lightgray!30}13.21 &	\cellcolor{lightgray!30}29.82 &	\cellcolor{lightgray!30}46.13
& \cellcolor{lightgray!30}7.40  & \cellcolor{lightgray!30}19.40 & \cellcolor{lightgray!30}37.87
& \cellcolor{lightgray!30}21.30 & \cellcolor{lightgray!30}45.59 & \cellcolor{lightgray!30}63.25\\ \midrule

\multirow{3}{*}{\parbox{0.07\textwidth}{\textsc{text code}}} 
& -
&\cellcolor{lightgray!30} 10.37 &	\cellcolor{lightgray!30}\textbf{15.66} &	\cellcolor{lightgray!30}\textbf{25.19}
& \cellcolor{lightgray!30}11.69 &	\cellcolor{lightgray!30}27.28 &	\cellcolor{lightgray!30}52.63
& \cellcolor{lightgray!30}\textbf{9.00} & \cellcolor{lightgray!30}22.46 & \cellcolor{lightgray!30}40.81
& \cellcolor{lightgray!30}20.23 & \cellcolor{lightgray!30}49.07 & \cellcolor{lightgray!30}\textbf{68.51}\\
&\textsc{partial}
& \textbf{11.59} &	14.96 &	23.83
& \textbf{16.66} &	32.32 &	51.75
& 7.80 & 22.63 & 41.70
& 23.08 & 49.74 & 68.12\\
&\textsc{full}
& \textbf{11.59} &	14.81 &	23.94
& 15.30 &	\textbf{33.36} &	\textbf{54.87}
& 7.80 & \textbf{22.69} & \textbf{41.77}& 
\textbf{23.39} & \textbf{50.77} & 68.22 \\ \arrayrulecolor{gray!30}\midrule

\multirow{3}{*}{\parbox{0.07\textwidth}{\textsc{Code}}}
& -
& 12.80 &	\cellcolor{blue!15}\textbf{16.81} &	26.94
& \cellcolor{blue!15}\textbf{17.98} &	34.54 &	54.61

& 8.80 & 23.29 & 41.46
& 23.81 & 52.37 & 70.10\\
&\textsc{partial}
& 12.80 &	16.47 &	26.25
& 17.13 &	32.83 &	52.42
& \textbf{9.00}& 23.28& \cellcolor{blue!15}\textbf{42.53}
& \textbf{24.44}& 52.35& 70.49\\
& full
& \cellcolor{blue!15}\textbf{13.41} &	15.97 &	\cellcolor{blue!15}\textbf{27.32}
& 17.88 &	\cellcolor{blue!15}\textbf{36.82} &	\textbf{57.91}
& 8.00 & \cellcolor{blue!15}\textbf{23.32} & 42.47
& 24.39 & \cellcolor{blue!15}\textbf{53.42} & \textbf{70.50}\\ \arrayrulecolor{gray!30}\midrule

\multirow{3}{*}{\parbox{0.07\textwidth}{\textsc{Corrupt Code}}}  
& - 
& \textbf{11.59} &	16.15 &	26.34
& 16.39 &	33.97 &	54.84
& 8.60 & \textbf{23.00}& \textbf{42.04}
& 22.46 & 51.36 & 69.60\\
& \textsc{partial} 
& \textbf{11.59} &	\textbf{15.96} &	27.04
& 16.24 &	33.45 &	54.07
& \cellcolor{blue!15}\textbf{9.20} & 22.59 & 41.36
& 23.57 & 50.56 & 68.32\\
& \textsc{full} 
& 10.98 &  15.73 &	\textbf{27.11}
& \textbf{17.51}	&  \textbf{36.67} &	\cellcolor{blue!15}\textbf{58.59}
& \cellcolor{blue!15}\textbf{9.20} & 22.92& 41.15
& \cellcolor{blue!15}\textbf{24.67} & \textbf{52.76}& \textbf{70.15} \\ \arrayrulecolor{gray!30}\midrule

\multirow{3}{*}{\parbox{0.07\textwidth}{\textsc{Prefix Code}}} 
&-
& 7.32 &	9.87 &	16.84
& 14.37 &	32.69 &	52.93
& 8.60& 20.93 & 39.89
& 23.63& 51.39 & 70.21\\
&\textsc{partial}
& 9.15 &	13.53 &	\textbf{22.18}
& 14.48 &	33.90 &	53.10
& \cellcolor{blue!15}\textbf{9.20} & 19.84 & 39.61 
& 24.19 & 50.42 & 68.02  \\
&\textsc{full} 
& \textbf{9.76} &	\textbf{13.67} &	22.00
& \textbf{16.30} &	\textbf{36.10} &	\textbf{56.56}

& \cellcolor{blue!15}\textbf{9.20} & \textbf{21.45}  & \textbf{40.59}
& \textbf{24.45} &  \textbf{53.10} & \cellcolor{blue!15}\textbf{70.72} \\\arrayrulecolor{black}\bottomrule
\end{tabular}
}
\caption{PyCodeGPT 100M model performance.}
\label{tab:pycodegpt_100m}
\end{subtable}

\vspace{0.2cm}

\begin{subtable}[b]{\linewidth}
\centering
\scalebox{0.75}{
    \begin{tabular}{lrrcc!{\color{gray!30}\vrule}ccc|ccc!{\color{gray!30}\vrule}rcc}
    \toprule
    & 
    & \multicolumn{3}{c}{\sc{\textbf{human eval}}}
    & \multicolumn{3}{c}{\sc{\textbf{INCR human eval}}}
    & \multicolumn{3}{c}{\sc{\textbf{mbpp}}}  
    & \multicolumn{3}{c}{\sc{\textbf{INCR mbpp}}} \\   
    & \textsc{separation} 
    &  \sc{p@$1$} &  \sc{p@$10$} &  \sc{p@$100$}  
    &  \sc{p@$1$} &  \sc{p@$10$} &  \sc{p@$100$}  
    &  \sc{p@$1$} &  \sc{p@$10$} &  \sc{p@$100$} 
    &  \sc{p@$1$} &  \sc{p@$10$} &  \sc{p@$100$} \\ 
    \cmidrule(lr){2-2} \cmidrule(lr){3-5} \cmidrule(lr){6-8} \cmidrule(lr){9-11} \cmidrule(lr){12-14}
           
\textsc{mapt}
& - 
& \cellcolor{lightgray!30}9.76	& \cellcolor{lightgray!30}17.07	& \cellcolor{lightgray!30}28.88
& \cellcolor{lightgray!30}20.70 & \cellcolor{lightgray!30}48.04 & \cellcolor{lightgray!30}63.08
& \cellcolor{lightgray!30}11.60 & 	\cellcolor{lightgray!30}24.99	 & \cellcolor{lightgray!30}44.84
& \cellcolor{lightgray!30}5.93  &   \cellcolor{lightgray!30}28.80	 & \cellcolor{lightgray!30}44.63

\\ \midrule

\multirow{3}{*}{\parbox{0.07\textwidth}{\textsc{Text Code}}}
& - 
& \cellcolor{lightgray!30}12.20 &	\cellcolor{lightgray!30}18.97 &	\cellcolor{lightgray!30}29.50
& \cellcolor{lightgray!30}22.93 &	\cellcolor{lightgray!30}49.96 &	\cellcolor{lightgray!30}64.23
& \cellcolor{lightgray!30}11.80 &	\cellcolor{lightgray!30}26.18 &	\cellcolor{lightgray!30}45.73
& \cellcolor{lightgray!30}11.78 &	\cellcolor{lightgray!30}38.99 &	\cellcolor{lightgray!30}61.38

\\
& \textsc{partial} 
& 11.59 &	\textbf{19.49} &	\textbf{31.70}
& 22.91 &	49.60 &	\textbf{64.91}
& \textbf{12.00} &	26.35 &	46.76
& 11.93 &	39.43 &	60.87

\\
& \textsc{full} 
& \cellcolor{blue!15}\textbf{13.41} &	19.31 &	29.61
& \textbf{23.54} &	\textbf{49.98} &	63.48
& 11.40 &	\textbf{26.65} &	\textbf{46.82}
& \textbf{12.80} &	\textbf{41.93} &	\textbf{63.27}
\\ \arrayrulecolor{gray!30}\midrule

\multirow{3}{*}{\parbox{0.07\textwidth}{\textsc{Code}}}
& -
& 12.20 &	19.63 &	\cellcolor{blue!15}\textbf{32.45}
& 23.65 &	50.25 &	64.89
& 12.20 &	26.69 & 46.61
& \textbf{14.76} &	\textbf{44.75} &	\textbf{64.84}

\\
& \textsc{partial}
& \cellcolor{blue!15}\textbf{13.41} &	\cellcolor{blue!15}\textbf{20.27} &	31.27
& \cellcolor{blue!15}\textbf{24.44} &	\cellcolor{blue!15}\textbf{50.63} &	65.46
& 12.00 &	\cellcolor{blue!15}\textbf{27.14} &	\cellcolor{blue!15}\textbf{47.08}
& 13.27 &	41.79 &	63.06

\\
& \textsc{full}
& 12.20 &	19.74 &	30.82
& 23.81 &	50.51 &	\cellcolor{blue!15}\textbf{66.21}
& \cellcolor{blue!15}\textbf{12.60} &	27.01 &	46.14
& 14.23 &	43.80 &	64.21
\\ \arrayrulecolor{gray!30}\midrule

\multirow{3}{*}{\parbox{0.07\textwidth}{\textsc{Corrupt Code}}}
& -
& 12.80 &	19.06 &	28.20
& 23.83 &	49.89 &	63.98
& \textbf{12.20} &	26.46 &	\textbf{46.57}
& 15.17 &	45.41 &	\cellcolor{blue!15}\textbf{65.95}

\\
& \textsc{partial}
& \cellcolor{blue!15}\textbf{13.41} &	\textbf{19.21} &	\textbf{29.70}
& 23.83 &	\textbf{50.12} &	\textbf{65.22}
& 11.80 &	26.44 &	45.15
& 13.53 &	42.36 &	62.85

\\
& \textsc{full}
& 12.80 &	18.65 &	28.54
& \textbf{23.91} &	49.48 &	64.02
& 11.60 &	\textbf{26.96} &	44.85
& \cellcolor{blue!15}\textbf{15.49} &	\cellcolor{blue!15}\textbf{46.38} &	65.60
\\ \arrayrulecolor{gray!30}\midrule

\multirow{3}{*}{\parbox{0.07\textwidth}{\textsc{Prefix Code}}}
& - 
& 10.98 &	18.33 &	27.56
& 23.27 &	49.98 &	63.78
& 11.00 &	26.10 &	45.90
& \textbf{11.23} &	\textbf{38.49} &	\textbf{60.80}

\\
& \textsc{partial}
& \textbf{12.80} &	19.26 &	\textbf{30.77}
& \textbf{23.88} &	\textbf{50.16} &	64.23
& 10.80 &	26.20 &	46.00
& 10.39 &	35.19 &	57.44

\\
& \textsc{full}
& \textbf{12.80} &	\textbf{19.37} &	30.21
& 23.59 &	49.64 &	\textbf{64.95}
& \textbf{11.20} &	\textbf{26.28} &	\textbf{46.27}
& 9.13 &	33.67 &	56.52
\\\arrayrulecolor{black}\bottomrule
\end{tabular}
}
\caption{PanGu-Coder 100M model performance.}
\label{tab:pangu_100m}
\end{subtable}

\caption{Pass@$k$ and Incremental pass@$k$ across objectives, embedding separation strategies and datasets. \textbf{Bold} numbers
denote best performance across separation methods and \colorbox{blue!15}{highlighted} denote overall best for each metric. 
}
\label{tab:results_100m}
\vspace{-3mm}
\end{table*}

\section{Zero-shot Results}

We first perform an ablation analysis on the 100M model variants to determine the gains based on our training data, the training objective and the embedding separation approaches.
Our primary comparisons are against two baseline model variants (marked with a gray background), i.e. after modality-agnostic (\textsc{mapt}) training 
and after standard next-token-prediction (\textsc{text-code}) on the entire sequence. The latter facilitates comparison against continual training on the given data for the same number of epochs as the other objectives. 

In Table~\ref{tab:pycodegpt_100m}, we evaluate PyCodeGPT across the two datasets and four objectives. Additional training with text-to-code pairs improves performance up to +1.2, +2.4 and +4.3 points for $k$=1,10,100 over \textsc{mapt} training on HumanEval and correspondingly +1.6, +3.0 and +2.9 on MBPP.
PyCodeGPT benefits more from the task-specific data as it was not exposed to them during \textsc{mapt}. 
Embedding separation additionally improves up to +3.0, +1.1, +2.1 pass@$k$ for HumanEval and up to +0.2, +0.8, +1.7 for MBPP. 
We observe stronger benefits when looking at incremental pass@$k$, with gains up to +6.2, +9.5, +5.9 on HumanEval and +4.4, +4.3, +2.2 on MBPP.

Moving on to Table~\ref{tab:pangu_100m} we show results for PanGu-Coder model of 100M parameters under the same datasets and methods. Firstly, we notice that we get an improvement of +2.4, +1.9 and +0.6 points for $k$=1,10,100 over \textsc{mapt} training on HumanEval and correspondingly +0.2, +1.1 and +0.8 on MBPP. 
We attribute this gain to the formulation of the data in a format that exactly matches the task at hand. 
Overall, embedding separation consistently offers additional improvements up to +1.2, +1.3, +2.9 on HumanEval and +0.8, +0.9, +1.3 on MBPP.
In terms of incremental pass@$k$, again we observe larger gains over \textsc{text-code}, mostly on MBPP with +3.7, +7.3 and +4.5 points.

Across training objectives, we notice that in the majority of settings, \textsc{code-clm} outperforms across the board. 
Across datasets, separation offers the most notable gains in the \textsc{prefix-code-clm} objective, probably because it helps the model to better adapt to bidirectional attention on the docstring. 
We note that most of the gains from embedding separation seem to affect pass@10 and pass@100, which implies that separation helps increase the expressiveness of the model but not the MAP (maximum a posteriori) solution as much.


\begin{table*}[t!]
\centering
\scalebox{0.75}{
\begin{tabular}{lrccc!{\color{gray!30}\vrule}ccc|ccc!{\color{gray!30}\vrule}rcc}   
 \toprule
    & 
    & \multicolumn{3}{c}{\sc{\textbf{human eval}}}
    & \multicolumn{3}{c}{\sc{\textbf{INCR human eval}}}
    & \multicolumn{3}{c}{\sc{\textbf{mbpp}}}  
    & \multicolumn{3}{c}{\sc{\textbf{INCR mbpp}}} \\   
    & \textsc{separation} 
    &  \sc{p@$1$} &  \sc{p@$10$} &  \sc{p@$100$}  
    &  \sc{p@$1$} &  \sc{p@$10$} &  \sc{p@$100$}
    &  \sc{p@$1$} &  \sc{p@$10$} &  \sc{p@$100$}
    &  \sc{p@$1$} &  \sc{p@$10$} &  \sc{p@$100$} \\ \cmidrule(lr){2-2} \cmidrule(lr){3-5} \cmidrule(lr){6-8} \cmidrule(lr){9-11} \cmidrule(lr){12-14}
    
\textsc{MAPT} & - 
& \cellcolor{lightgray!30}16.46 &	\cellcolor{lightgray!30}24.51 &	\cellcolor{lightgray!30}35.39
& \cellcolor{lightgray!30}27.82 &	\cellcolor{lightgray!30}55.68 &	\cellcolor{lightgray!30}68.34
& \cellcolor{lightgray!30}18.80 & \cellcolor{lightgray!30}35.36 & \cellcolor{lightgray!30}53.24
& \cellcolor{lightgray!30}8.14 & \cellcolor{lightgray!30}35.26 & \cellcolor{lightgray!30}57.77\\\midrule

\parbox{0.07\textwidth}{\multirow{3}{*}{\textsc{Text Code}}}
& - 
& \cellcolor{lightgray!30}\textbf{18.90} &	\cellcolor{lightgray!30}25.79 &	\cellcolor{lightgray!30}37.46
& \cellcolor{lightgray!30}\textbf{31.39} &	\cellcolor{lightgray!30}56.36 &	\cellcolor{lightgray!30}69.22
& \cellcolor{lightgray!30}17.60 &	\cellcolor{lightgray!30}37.59 &	\cellcolor{lightgray!30}54.37
& \cellcolor{lightgray!30}\textbf{10.95} &	\cellcolor{lightgray!30}\textbf{42.27} &	\cellcolor{lightgray!30}\textbf{66.62}\\
 
& 
partial
& 18.29 &	25.54 &	37.82
& 30.81 &	\textbf{56.39} &	\textbf{69.44}
& \textbf{18.00} &	\cellcolor{blue!15}\textbf{37.70} &	\cellcolor{blue!15}\textbf{55.60}
& 9.97 &	39.19 &	64.93\\

& full
& 18.29 &	\textbf{26.48} &	\cellcolor{blue!15}\textbf{41.30}
& 30.85 &	56.28 &	68.81
& 16.40 &	37.43 &	55.35
& 9.71 &	40.22 &	65.49\\ \midrule

\multirow{3}{*}{\textsc{Code}} 

& 
-
& 20.73 &	25.96 &	37.79
& 32.16 &	56.66 &	69.88
& \cellcolor{blue!15}\textbf{18.80} &	\textbf{37.54} &	\textbf{54.99}
& 12.27 &	44.87 &	\cellcolor{blue!15}\textbf{68.65}\\

& 
partial
& \cellcolor{blue!15}\textbf{21.34} &	26.27 &	37.94
& \cellcolor{blue!15}\textbf{32.88} &	56.82 &	69.20
& 18.60 &	37.36 &	54.39
& \cellcolor{blue!15}\textbf{12.38} &	\cellcolor{blue!15}\textbf{45.37} &	68.58\\ 

& full
& 18.29 &	\cellcolor{blue!15}\textbf{26.98} & \textbf{39.94}
& 30.10 &	\cellcolor{blue!15}\textbf{57.12} &	\cellcolor{blue!15}\textbf{70.15}
& \cellcolor{blue!15}\textbf{18.80} &	37.45 &	54.33
& 9.16 &	37.84 &	63.13\\\bottomrule

\end{tabular}
}
\caption{PanGu-Coder 350M CodeLM performance.}
\label{tab:350m_results}
\vspace{-3mm}
\end{table*}

\subsection{Scaling Up}

To determine whether our observations regarding embedding separation hold for different model sizes, we also
test our hypothesis on a larger version of PanGu-Coder, consisting of 350M parameters, for the reported best training objective \textsc{code-clm} of its smaller variant (see Table~\ref{tab:results_100m}).
In Table~\ref{tab:350m_results}, we observe again improvements over MAPT with +2.4, +1.2, +2.0 pass@1/10/100 on HumanEval.
Similar trends can be observed on the MBPP dataset, though the differences appear smaller due to its larger sample size and difficulty.
More evident improvements are noticed over no separation with +2.4, +1.9, +3.8 on HumanEval and +1.2, +0.1, +1.2 on MBPP, respectively. 
Incremental pass@$k$ shows +1.4, +0.7, +0.9 and +1.4, +3.1, +2.0 over each dataset. We present a comparison against other Code LMs in Appendix~\ref{app:relatedworkcomparison}.

\section{Visualising Separated Embeddings}

\begin{figure}[t!]
    \centering
    \includegraphics[width=\linewidth]{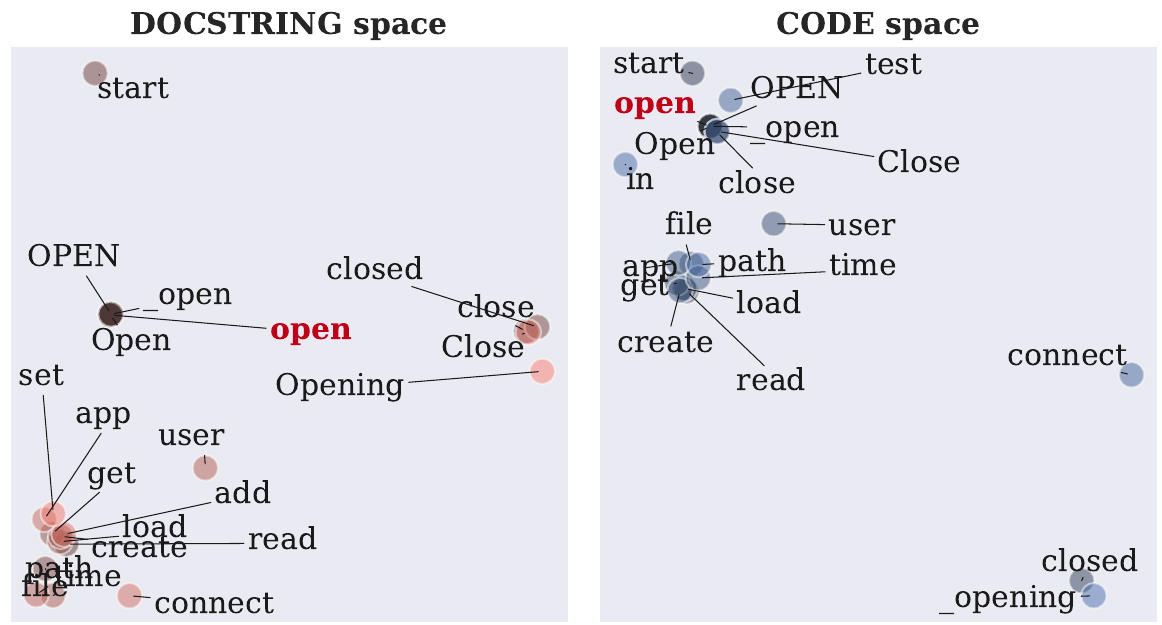}
    \includegraphics[width=\linewidth]{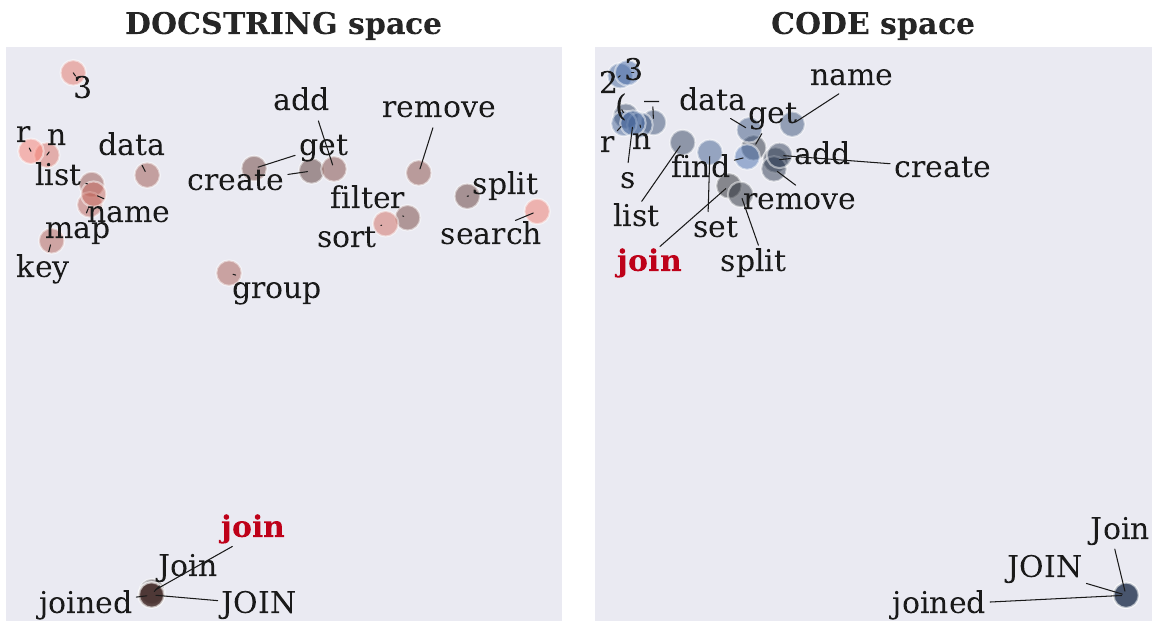}
 \caption{T-SNE plots of the top 20 neighbours of the tokens \texttt{open} (top) and \texttt{join} (bottom) on different embedding spaces for PanGu-Coder 350M parameter model with partial embedding space separation (PES).}
\label{fig:node_codeclm_tsne}
\end{figure}

We attempt to qualitatively check the embedding space separation by visualising token embeddings on a 2D plane via T-SNE plots. 
Figure~\ref{fig:node_codeclm_tsne} illustrates the 
20 closest neighbours of the tokens \texttt{open} and \texttt{join} in the two modality spaces, as measured using cosine similarity for PanGu-Coder 350M model with partial embedding separation (PES).

For \texttt{open}, in the docstring space we observe that its representation is separate from \texttt{close} or other build-in operations, e.g. \texttt{get}. On the contrary, in the code space \texttt{open} and \texttt{close} are grouped together with other operations that are used in a similar way in code. 
Analogously, \texttt{join} in the docstring space is close to words with a similar natural language meaning and surface form, while in the code space its representation is close to function-alike tokens in Python such as \texttt{remove}, \texttt{get}, \texttt{split}, etc.
We observe similar behaviours for other tokens; see additional plots in Appendix~\ref{app:extra_tsnes}.

\begin{table}[t!]
    \centering
    \scalebox{0.75}{
    \begin{tabular}{lccc}
        \toprule
        \textsc{Test Set} 
        & \textsc{HumanEval} & \textsc{MBPP Dev} & \textsc{MBPP Test}\\  
       \cmidrule(r){1-1} \cmidrule(l){2-4}
       \textsc{HumanEval} &  -  & 0.099	& 0.163\\  
       \textsc{MBPP Dev} & -	& -	& 0.046  \\ 
       \textsc{MBPP Test} & -	& -	& - \\ 
       \bottomrule
    \end{tabular}
    }
    \caption{Average Pearson correlation between pass@1 performance of different model checkpoints.}
    \label{tab:datasetgeneralization}
\end{table}

\section{Generalisation Analysis and Discussion}

We performed an analysis over the performance of all the checkpoints of our trained models (checkpoints were cached every 10K steps), to determine whether performance on any test set was predictive on the performance of other sets; we included HumanEval and both development and test sets of MBPP. Pearson analysis between checkpoint performance suggests very weak correlation between model performance overall; consult Table~\ref{tab:datasetgeneralization}.

This seems to indicate that the distribution of problems that are covered by any one test set is relatively unique, and raises the question whether comparisons over any particular dataset should be expected to generalise on unseen problems when models are deployed. Potentially, this is partially explained by the small size of the test sets, but could be indicative of a larger problem; that due to the specificity of problems contained in these test sets, they do not evaluate generalisable model coding capabilities as much as they measure whether they have been exposed to very similar problems and definitions during pre-training. To offer an example, if the model has not been exposed to problems regarding Fibonacci sequences during pre-training, it is unrealistic to expect it to generate a relevant code solution in a zero-shot setting. Similarly, whether a model can generate a solution to a Fibonacci problem offers no indication of the its capabilities to solve other problems.

\section{Related Work}

Recently, there has been an increasing interest in extending proven natural language understanding and generation methods to code understanding and generation tasks.
CodeBERT~\cite{feng-etal-2020-codebert}, for instance, was trained using a combination of Masked Language Modelling inspired by \citet[BERT]{devlin-etal-2019-bert} and Replaced Token Detection from \citet[ELECTRA]{clark-etal-2020-electra}. CodeT5~\cite{wang-etal-2021-codet5} and PYMT5~\cite{clement-etal-2020-pymt5} were built on top of \cite[T5]{raffel-etal-2020-exploring} while UniXcoder~\cite{guo-etal-2022-unixcoder} was based on UniLM~\cite{dong-etal-2019-unilm} and combines three pre-training objectives using different attention masks to control access to surrounding context for a token to be predicted. 
Focused on the task of text-to-code generation, \citet{codex} introduced CodeX, a set of GPT-based language models trained on publicly available code from GitHub, up to 12B parameters in size. 
\citet{alpha_code} introduced AlphaCode, a set of sequence-to-sequence models with up to 41B parameters, trained on data from programming competitions, e.g. Codeforces\footnote{\url{https://codeforces.com/}} as well as GitHub code in several programming languages. 
CodeGen~\cite{nijkamp2023codegen} was proposed as a conversational text-to-code approach using large language models with sizes of up to 16B parameters. 
The authors proposed three model variants, trained on The Pile~\cite{thepile}, 
continuously trained on BigQuery (with 6 programming languages) and finally trained on Python-only data.
InCoder \cite{fried2022incoder} extends left-to-right code generation with an infilling training objective, similar to \citet{bavarian2022efficient}, and is able to predict spans of partial programs as well. 
SantaCoder~\citep{allal2023santacoder} is one of the latest Code LMs, sizing up to 1.1B parameters, supporting Python, Java, and JavaScript. It was trained with multi-query attention and the fill-in-the-middle (FIM) objective~\citep{bavarian2022efficient} on The Stack~\citep{kocetkov2022stack}, a 3 TB publicly available dataset supporting 358 programming languages collected from permissively-licensed source code files from Github. StarCoder~\citep{li2023starcoder} is a 15.5B parameter model, similarly pretrained with multi-query attention and FIM on 80+ programming languages from The Stack. 
With the exception of CodeX \cite{codex} calculating CLM loss on code exclusively, this is the first work to consider modality-relative embedding separation and training objectives. 

Methods that consider NL and code as different modalities mostly focus on taking into account different \textit{views} of code. For instance, GraphCodeBERT~\cite{guo2021graphcodebert}, noted that previous pretrained models treat code snippets as sequences of tokens while ignoring the inherent structure of code. They presented GraphCodeBERT, which showed that incorporating the \textit{data flow}, i.e. a semantic-level structure of code extracted from the Abstract Syntax Tree (AST), leads to richer code representations. \citet[TREEBert]{jiang2021treebert} instead, used the actual AST together with code snippets.
To the best of our knowledge we are not aware of any work assigning different embeddings to tokens of the same sequence depending on which type of text (natural language/code) they appear.

\section{Conclusion}

Existing CodeLMs consider both code and natural language as a single modality, mapping them into a shared embedding space. 
However, in this work, we posit that the semantics and usage of tokens can differ between code and NL, requesting for a possible space separation.
As such, we proposed to consider code and natural language as different modalities for the task of text-to-code generation by introducing modality-aware embedding separation strategies and training objectives.
In detail, assuming a general CodeLM trained on raw data--where modalities are shared--we continue training on task-specific data with separated embedding spaces.
We present partial separation, which targets language-specific tokens, as well as full separation that duplicates the entire model's vocabulary.
In addition, we proposed incremental pass@$k$, as a variant of the standard metric that evaluates the code completion capabilities of models.

Zero-shot evaluation on the HumanEval and MBPP datasets, with two 100M and one 350M parameters models, indicate that embedding space separation improves code generation across different objectives. We also observe that further pre-training on data formatted to match the target task consistently boosts performance.

Although there is no clear winner among partial versus full space separation, each method comes with advantages and drawbacks. While partial separation minimally increases the model's vocabulary, it requires manual effort for each target programming language. On the other end, full separation is automatic but significantly increases vocabulary size.
Given these observations, future work could target efficient approaches for separating only certain tokens to strike a balance between the two.



\section*{Limitations}

We enumerate the limitations of this work as part of our experimental process. Firstly, our analysis was focused on Python only. While our proposed methods are orthogonal to the programming language they are applied to, it remains to be confirmed that our findings generalise to other languages. Overall, our observations should stand for high-level programming languages that resemble natural language.
Secondly, our trained models are limited to function-level text-to-code generation (both program synthesis and code completion), as we opted for a particular use case to study the connection between programming and natural language. These models are unable to perform multi-turn generation or generation of multiple code functions given a problem description, as mentioned in Section~\ref{sec:evaluation}. We leave such explorations for future work.

\section*{Acknowledgements}
The authors would like to thank Huawei Cloud for collecting the Python data. We also acknowledge the MindSpore team for providing technical support.\footnote{\url{https://www.mindspore.cn/en}}

\bibliography{anthology,custom}

\appendix

\section{Data Collection and Processing}
\label{sec:data}

\subsection{Collection}

We crawled existing, public repositories from GitHub before May 2021, resulting in approximately 65 million Python files with a total size of 380 GB. We then removed duplicate files based on the rowKey of each file's MD5, which resulted in 40 million files (186 GB). 
We further kept files that meet the following criteria:
(a) the file size is under 1MB; (b) the code is Python3 compatible, using Abstract Syntactic Tree (AST) parsing; (c) there are fewer than 100 characters per line on average; (d) and there are fewer than $1,000$ characters in any single line.
We then removed the duplicated functions from the remaining files. In the end, 120 GB of training data was obtained.

\subsection{Pre-processing} 

The collected data include both plain Python code and code accompanied by a natural language description after the function signature.
The latter can be used to create text-to-code instances which are considered task-specific data for program synthesis, while code-only function snippets can expose models to generic Python programming.

We apply AST parsing\footnote{\url{https://tree-sitter.github.io/tree-sitter/}} on the remaining Python files to extract valid functions and their corresponding docstrings.\footnote{The strings that follow Python docstring conventions: \url{https://peps.python.org/pep-0257/}} While we also extract classes, we will only refer to functions as the process is identical. An example is shown below:
\begin{lstlisting}[style=python]
def gcd(a: int, b: int): -> int
    """
    Return a greatest common divisor of the two integers a and b
    """
    while b:
        a, b = b, a % b
    return a
\end{lstlisting}
For all extracted functions, we also remove comments. Many comments are simple housekeeping messages, such as dates, contributor's identity and ``Do not delete''.\footnote{We leave training with comments as future work.}
We then replace new lines, indentation and dedentation with \texttt{[new\_line]}, \texttt{[indent]} and \texttt{[dedent]}, respectively, to normalise spaces, which effectively reduces the length of input sequences to the model.

Finally, we apply deduplication to all available training data. In the end, 
we gather 25 million text-to-code pairs and 80.8 million code-only function snippets, respectively.

\section{Training of PanGu-Coder}
\label{app:pangu_mapt}

Here we provide additional details with respect to PanGu-Coder's modality-agnostic pre-training.
\subsection{Tokenization}
\label{sec:tokenization}

We use SentencePiece~\cite{kudo-richardson-2018-sentencepiece} as our primary tokenizer. 
We train the tokenizer on 10 million samples randomly drawn from the training data, including both docstrings and code. 
During training, the normalising symbols, \texttt{[new\_line]}, \texttt{[indent]} and \texttt{[dedent]}, are passed to SentencePiece to preserve their integrity.
The tokenizer's vocab size is set to $32,000$ tokens.

\subsection{Data Formatting}

\begin{figure}[t!]
    \centering
    \begin{subfigure}[b]{\linewidth}
    \centering
        \includegraphics[width=0.95\linewidth]{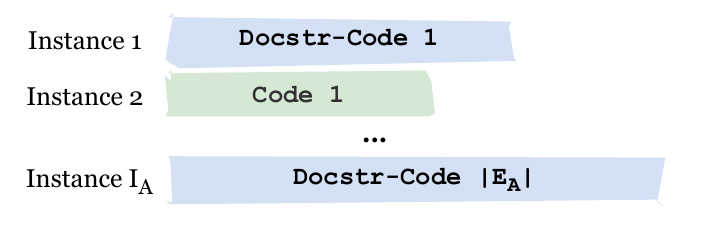}
        \caption{MAPT samples, using all available data.}
        \label{fig:stage1_data}
    \end{subfigure}
    
    \begin{subfigure}[b]{\linewidth}
    \centering
        \includegraphics[width=0.95\linewidth]{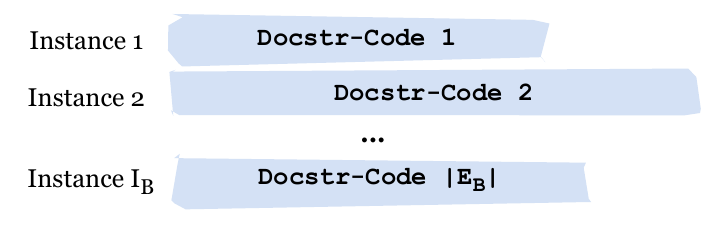}
        \caption{MRPT samples, using only docstring-code pairs.}
        \label{fig:stage2_data}
    \end{subfigure}
    \caption{Input data formats during modality-agnostic (MAPT) and modality-relative pre-training (MRPT).}
\end{figure}

When a docstring exists within a function, we form a text-to-code instance by re-organising the function according to the following template, \texttt{[descr] \textit{docstring} [python] \textit{code} [eoc]}, where \texttt{[descr]} stands for the beginning of a problem description placeholder, \texttt{[python]} for start of python code and \texttt{[eoc]} for end-of-code, as follows:
\begin{lstlisting}[style=custom]
[descr] Return a greatest common divisor of two integers a and b [python] def gcd(a: int, b: int) -> int: [new_line] [indent] while b: [new_line] [indent] a, b = b, a % b [new_line] [dedent] return a [eoc]
\end{lstlisting}
In case there is no available docstring for a code snippet we simply omit the \texttt{[descr]} portion. 
In our MAPT training, we use all the available training samples including text-to-code pairs and code-only instances.
For MRPT training, we consider only text-to-code instances, as shown in Figure~\ref{fig:stage2_data}.

\subsection{Sample Concatenation}
We observe that a large portion of the training samples are shorter than $512$ tokens after tokenization. Padding the samples up to the context size can increase training time and waste resources. In this work, we adopt a data concatenation approach, known as packing, to improve training and energy efficiency.
We shuffle the data and start appending examples until the maximum sequence length is reached, forming a new sample. If appending an instance exceeds the maximum sequence length, then we continue forming the next sample.
Although padding is not completely eliminated with this technique, the amount of padding is greatly reduced and we end up with only 23.2M concatenated training samples 
almost 18\% of the original training samples for Modality-agnostic training of PanGu-Coder. Similarly, for Modality-relative training we have 6.1M examples for PyCodeGPT (26\% of original) and 5.7M for PanGu-Coder (24\% of original).

Since each concatenated training sample contains several original training samples, we use the \texttt{[eoc]} placeholder as an anchor to reset the attention mask and position ids, so that each unique sample only attends to itself.

\section{Additional Training Details}
\label{app:training_details}

We report the size of the models that we trained in terms of model settings in Table~\ref{tab:model_sizes}.
We used 8 V100 32GB GPUs for training the 100M models and 16 to train the 350M ones. The total training time of PyCodeGPT was approximately 5 days, 7 days for PanGu-Coder 100M model and 10 days for the 350M model.

\begin{table*}
    \centering
    \scalebox{0.8}{
    \begin{tabular}{lrrrrrr}
    \toprule
Model & Model Params & Layers & FFN size & Heads & Context Size & Vocab \\
& & & & & ($c_\textsc{cntx}$) & ($n_\textsc{vocab}$) \\
        \midrule
        PyCodeGPT~\citep{cert2022zan} & 111 M & 12 & 768 & 12 & 1,024 & 32,000 \\
        PanGu-Coder~\citep{christopoulou2022pangu} & 118 M & 12 & 768 & 12 & 1,024 & 32,012 \\
        PanGu-Coder~\citep{christopoulou2022pangu} & 349 M & 24 & 1,024 & 16 & 1,024 & 32,012 \\
        \bottomrule
    \end{tabular}
    }
    \caption{Models configuration.} 
\label{tab:model_sizes}
\end{table*}

\section{Comparison with other Code PLMs}
\label{app:relatedworkcomparison}

We compare our best setting (i.e. PanGu-Coder with partial embedding separation) with existing Code LMs in Table~\ref{tab:human_eval_results}, reporting model and data sizes, the contextual window allowed ($c_\textsc{cntx}$), vocabulary size ($n_\textsc{vocab}$) and number of tokens the models were trained for. 

For \textsc{Codex}~\citep{codex}, the total data size and the number of trained tokens are calculated by considering the initial training of GPT-3~\cite{brown2020language} for 300 billion tokens on a collection of data equivalent to 570GB of compressed plain text. 
We include the decoder-only baseline presented by \textsc{AlphaCode}, and not the encoder-decoder model, as HumanEval results are only reported on the former. The number of train tokens of this baseline is not reported.
For \textsc{CodeGen}~\citep{nijkamp2023codegen} models, the dataset size of \textsc{CodeGen-Multi} was computed by accumulating The Pile~\cite{thepile} and BigQuery\footnote{\url{https://cloud.google.com/bigquery/public-data/}}, while \textsc{CodeGen-Mono} was additionally trained on BigPython. To calculate the number of training tokens for the \textsc{CodeGen} models, we assume the batch size reported in the paper corresponds to tokens instead of instances.
For \textsc{InCoder}, the vocabulary size was calculated as 55\% of GPT-2 vocabulary, based on \citet{fried2022incoder}.
For \textsc{SantaCoder} the reported numbers and the number of trained tokens were collected from the model card on the HuggingFace Hub.\footnote{\url{https://huggingface.co/bigcode/santacoder}} 
For the other models, explicit information was provided in the corresponding papers.

\begin{table*}[t!]
    \centering
    \scalebox{0.73}{
    \begin{tabular}{lrcrrrrrrrrr} 
    \toprule
        \multirow{2}{*}{\textsc{Model}} 
        & \multirow{2}{*}{\textsc{size}} 
        & \multirow{2}{*}{$n_\textsc{cntx}$}
        & \multirow{2}{*}{$n_\textsc{vocab}$}
        & \multicolumn{1}{c}{\textsc{data}}
        & \multicolumn{1}{c}{\textsc{train}} 
        & \multicolumn{3}{c}{\textbf{\textsc{HumanEval}} (\%)}  
        & \multicolumn{3}{c}{\textbf{\textsc{MBPP}} (\%)}\\ 
        &  &  & & \multicolumn{1}{c}{\textsc{(gb)}} & \multicolumn{1}{c}{\textsc{tokens}}
        & \textsc{p@$1$} & \textsc{p@$10$} & \textsc{p@$100$}
        & \textsc{p@$1$} & \textsc{p@$10$} & \textsc{p@$100$}
        \\   
        \arrayrulecolor{black}\cmidrule(r){1-6} \cmidrule(r){7-9} \cmidrule(r){10-12}

        \rowcolor{lightgray!20}
        \multirow{1}{*}{\textbf{\textsc{PanGu-Coder (cclm-partial)}}}
        & 300 M & 1,024 & 32.4 K & 97 & 179 B 
        & \textbf{21.3} & \textbf{26.3} & \textbf{37.9s} 
        & \textbf{18.6}	& 37.4 & 54.4	\\
        
        \multirow{1}{*}{\textsc{Codex}}
        & 300 M  & 4,096 & 50 K & 729 & 400 B 
        & 13.2 & 20.4 & 36.3 
        & -& - & -   \\ 
        
        \multirow{1}{*}{\textsc{AlphaCode}}
        & 302 M & 2,304  & 8 K &  715 & - 
        & 11.6 & 18.8  & 31.8 
        & - & - & -  \\
        
        \multirow{1}{*}{\textsc{CodeGen Multi} }
        & 350 M &  2,048 & 50 K & 1,595  & 250 B  
        &  6.7 & 10.6 & 16.8 
        &  7.5 & 24.2 & 46.4 \\
        
        \multirow{1}{*}{\textsc{CodeGen Mono}}
        & 350 M &  2,048 & 50 K & 1,812 & 325 B  
        & 12.8 & 23.1 & 35.2 
        & 14.6 &  \textbf{41.5} &  \textbf{63.0}\\
        
        \multirow{1}{*}{\textsc{PanGu-Coder}}
        & 317 M & 1,024  &  42 K &  147 &  211 B 
        &  17.1 &  24.1 &  34.6 
        & 16.2 & 34.4 & 53.7 \\
        
        \midrule
    
        \multirow{1}{*}{\textsc{AlphaCode}}
        & 1.1 B & 2,304  & 8 K &  715 & - 
        & 17.1 & 28.2  & 45.3  
        & - & - & - \\
        \multirow{1}{*}{\textsc{SantaCoder}} 
        & 1.1 B & 2,048 & 49 K &  268 & 236 B 
        & 18.0 & 29.0 & 49.0 
        & 35.0 & 58.0 & 77.0 \\
        

        \multirow{1}{*}{\textsc{InCoder}}  
        & 1.3 B & 2,048 & 27.6 K & 204 & 52 B 
        & 8.0 & - & - 
        & 10.9 & - & - \\ 
        \midrule

        \multirow{1}{*}{\textsc{InCoder}}  
        & 6.7 B & 2,048 & 27.6 K & 204 & 52 B 
        & 15.2 & 27.8 & 47.0 
        & 19.4 & - & -  \\
        
        \bottomrule
    \end{tabular}
    }
    \caption{Pass@$k$ rates on the HumanEval dataset, among various models. Sizes are reported in thousands (K), millions (M), billions (B) and trillions (T). CB refers to CodeBLUE.
    Models: CodeX~\cite{codex}, AlphaCode~\cite{alpha_code}, CodeGen~\cite{nijkamp2023codegen}, PanGu-Coder~\cite{christopoulou2022pangu}, SantaCoder~\cite{allal2023santacoder}, InCoder~\cite{fried2022incoder}\protect\footnotemark. Best results across the 300M models are \textbf{bold}ed.}
    \label{tab:human_eval_results}
\end{table*}

For all models, pass@$k$ rates are computed with $200$ samples, except for \textsc{AlphaCode} where the reported rates used $1,000$ samples and \textsc{CodeGen} which used $100$ samples.
Our proposed model achieves the best performance among similar-sized models across all metrics, even having been trained on much less training data\footnotetext{We did not include even larger scale models, since they would not be directly comparable with this work.}.

Regarding the MBPP dataset, 
our model outperforms all other reported models on pass@$1$, while being second best for pass@$10$ and pass@$100$. It also outperforms \textsc{InCoder} despite being significantly smaller.
Overall, we observe, that our PanGu-Coder performs better for lower $k$ values. We speculate that the gap in performance for pass@$100$ is a result of the small context size ($1,024$) the model has been trained on, which prevents the model from learning to generate long solutions, or solutions that have to attend over quite long descriptions. 
Allowing the model to be trained on longer inputs can be beneficial when scaling to many more sample solutions, which enables the generation of exhaustive solutions, e.g. enumerating all possible cases of a for loop.

Finally, the numbers of SantaCoder are using the MultiPL-E benchmark~\cite{cassano2022scalable} that includes small changes compared to the original datasets (e.g. three problems were removed from the HumanEval test set). As such, aside from the difference in model size, performance is not directly comparable.

\section{Additional T-SNE plots}
\label{app:extra_tsnes}

\begin{figure*}[t!]
    \centering
    \includegraphics[width=0.47\linewidth]{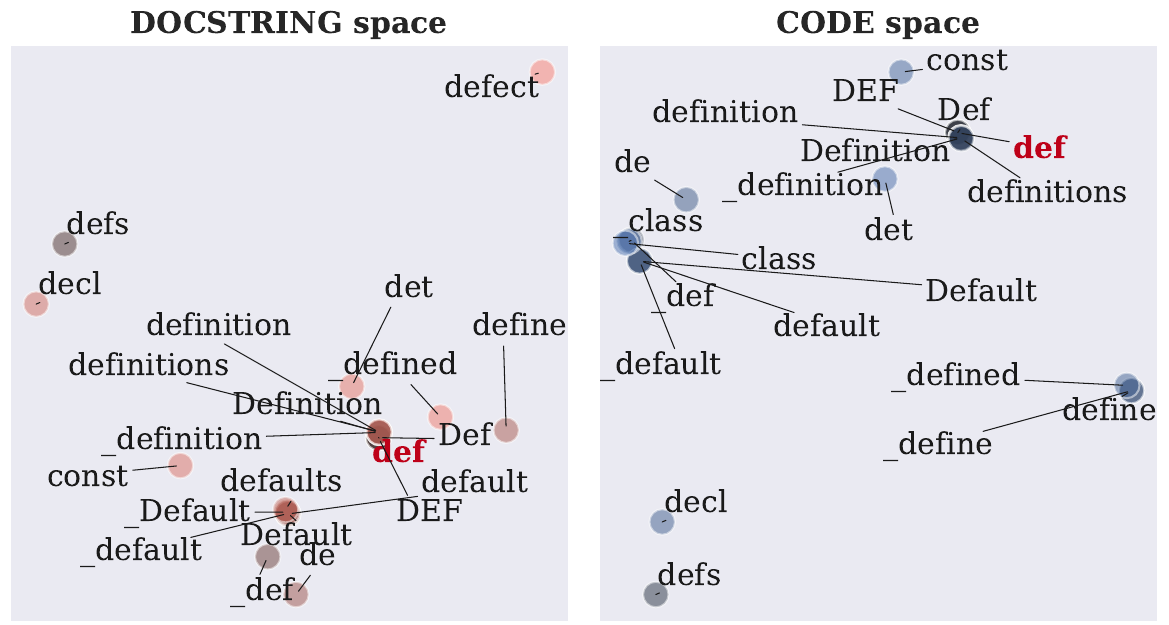}
    \includegraphics[width=0.47\linewidth]{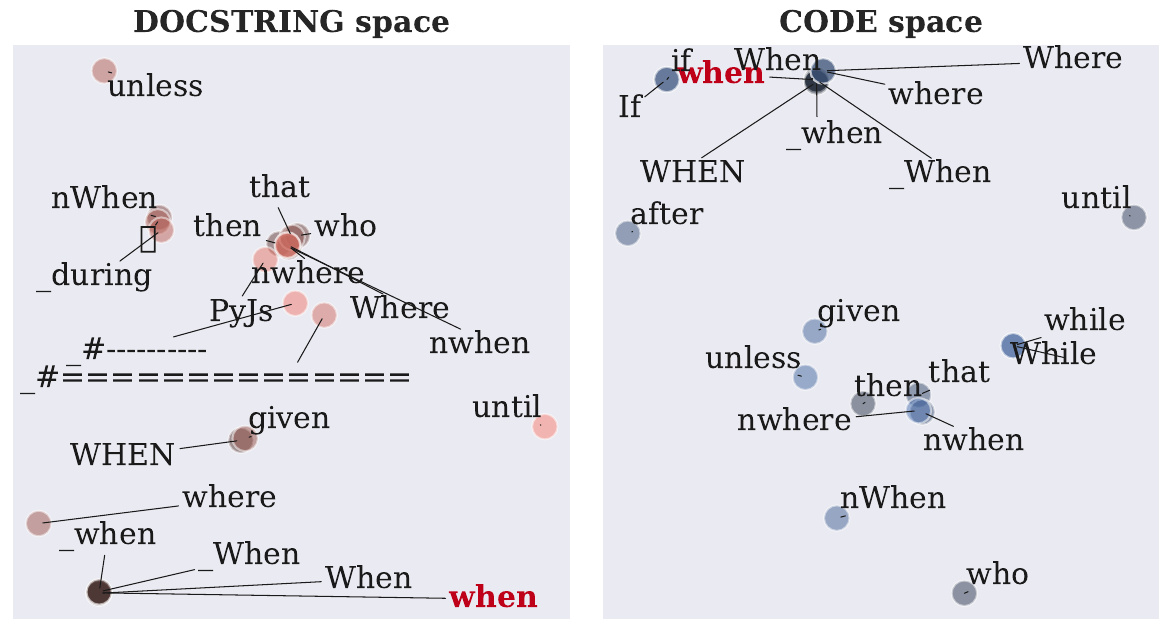}
    
    \includegraphics[width=0.47\linewidth]{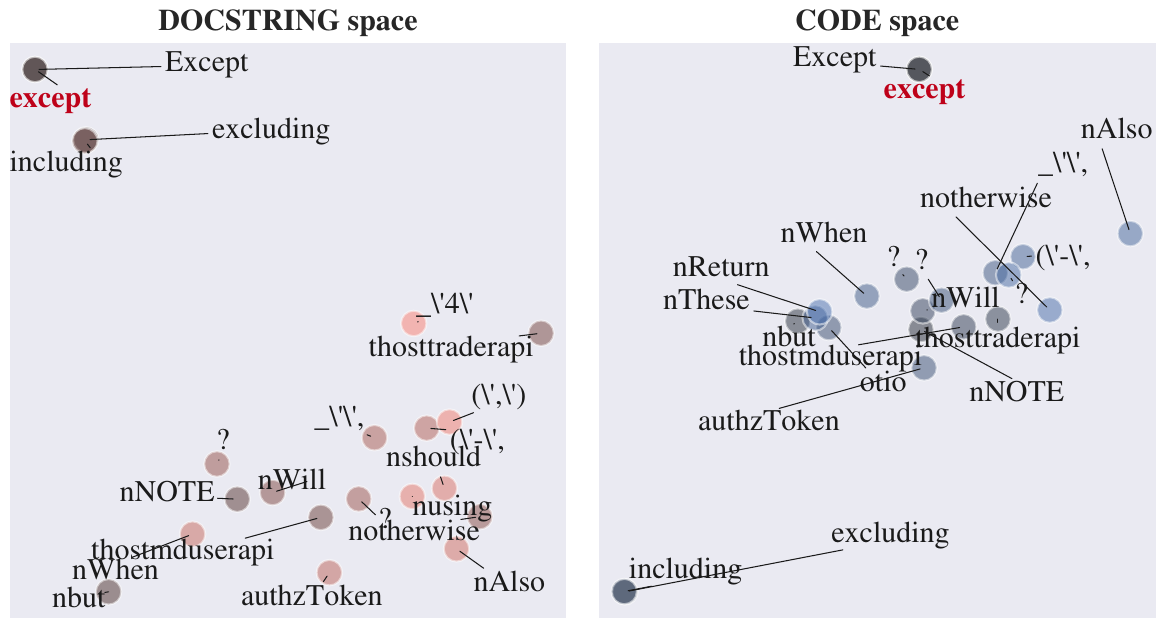}
    \includegraphics[width=0.47\linewidth]{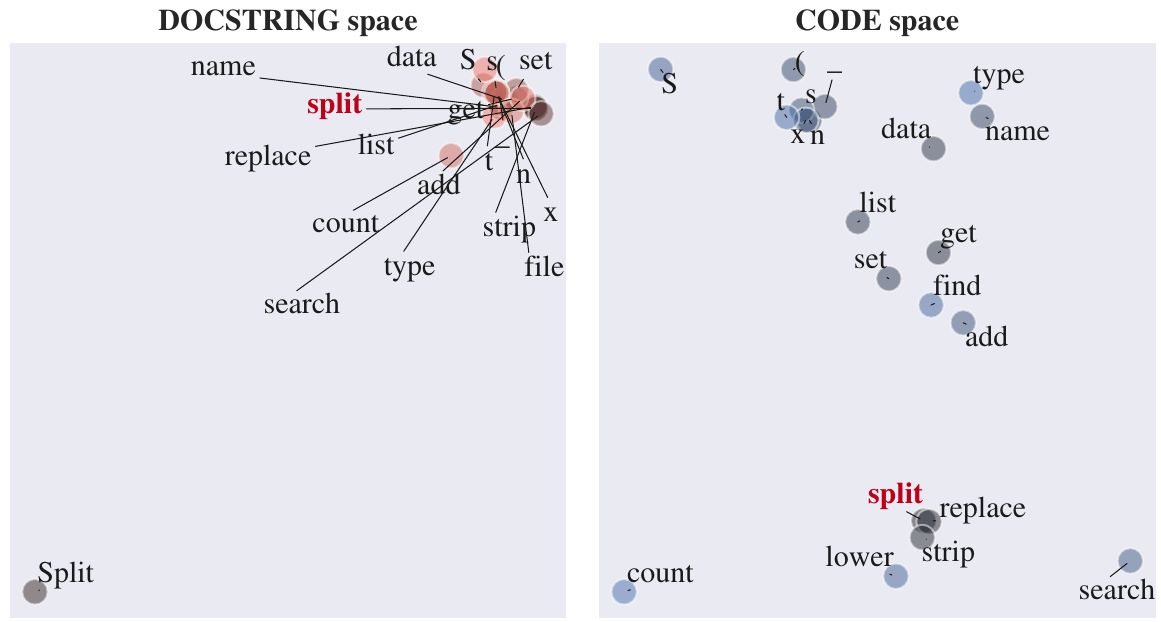}
    \caption{T-SNE plots of the top 20 neighbours of the tokens \texttt{def}, \texttt{when}, \texttt{except} and \texttt{split} (top to bottom) on different embedding spaces.}
    \label{fig:extra_tsnes}
\end{figure*}

We show additional T-SNE plots in Figure \ref{fig:extra_tsnes} as part of a qualitative analysis of the embedding separation between modalities for the \textsc{code-clm} objective and the 350M PanGu-Coder model.\footnote{We attribute credits to \url{https://github.com/Phlya/adjustText} for the aligning labels with points.}

In the scenario of \texttt{def}, we can see that \textit{definition} go further apart from \textit{default} as we move to the code embedding space.
For \texttt{when}, the neighbours in each space are quite different. In the docstring space, some random tokens appear that are not present in the code space.
For \texttt{except}, \textit{including} and \textit{excluding} in NL are close in the docstring space but further apart in the code space.
Finally, for \texttt{split}, we observe it gets closer with \texttt{replace}, \texttt{lower} and \texttt{strip} in the code space while in the docstring space, it can be found bundled together with many other tokens.

\end{document}